\documentclass{article}
\PassOptionsToPackage{numbers,compress}{natbib}
\usepackage[preprint]{neurips_2020}
\usepackage[utf8]{inputenc}
\usepackage[T1]{fontenc}
\usepackage{url}
\usepackage{booktabs}
\usepackage{amsfonts}
\usepackage{nicefrac}
\usepackage{microtype}

\usepackage{algorithm}
\usepackage{algorithmicx}
\usepackage[noend]{algpseudocode}
\usepackage{amsmath}
\usepackage{amssymb}
\usepackage{amsthm}
\usepackage{bbm}
\usepackage{bm}
\usepackage{color}
\usepackage{dirtytalk}
\usepackage{dsfont}
\usepackage{enumerate}
\usepackage{graphicx}
\usepackage{mathtools}
\usepackage{subfigure}
\usepackage{times}
\usepackage{wrapfig}
\usepackage{xspace}

\usepackage[usenames,dvipsnames]{xcolor}
\usepackage[bookmarks=false]{hyperref}
\hypersetup{
  pdftex,
  pdffitwindow=true,
  pdfstartview={FitH},
  pdfnewwindow=true,
  colorlinks,
  linktocpage=true,
  linkcolor=Green,
  urlcolor=Green,
  citecolor=Green
}
\usepackage[capitalize,noabbrev]{cleveref}

\usepackage[backgroundcolor=white]{todonotes}

\newcommand{\commentout}[1]{}
\newcommand{\junk}[1]{}

\Crefname{corollary}{Corollary}{Corollaries}
\Crefname{proposition}{Proposition}{Propositions}
\Crefname{theorem}{Theorem}{Theorems}
\Crefname{definition}{Definition}{Definitions}
\Crefname{assumption}{Assumption}{Assumptions}
\Crefname{example}{Example}{Examples}
\Crefname{remark}{Remark}{Remarks}
\Crefname{setting}{Setting}{Settings}
\Crefname{lemma}{Lemma}{Lemmas}

\usepackage{thmtools}
\declaretheorem[name=Theorem,refname={Theorem,Theorems},Refname={Theorem,Theorems}]{theorem}
\declaretheorem[name=Lemma,refname={Lemma,Lemmas},Refname={Lemma,Lemmas},sibling=theorem]{lemma}

\declaretheorem[name=Assumption,refname={Assumption,Assumptions},Refname={Assumption,Assumptions}]{assumption}

\newcommand{\cN}{\mathcal{N}}
\newcommand{\cP}{\mathcal{P}}

\newcommand{\eps}{\varepsilon}

\newcommand{\realset}{\mathbb{R}}

\newcommand{\E}[1]{\mathbb{E} \left[#1\right]}
\newcommand{\condE}[2]{\mathbb{E} \left[#1 \,\middle|\, #2\right]}

\newcommand{\prob}[1]{\mathbb{P} \left(#1\right)}

\newcommand{\var}[1]{\mathrm{var} \left[#1\right]}

\newcommand{\abs}[1]{\left|#1\right|}
\newcommand{\ceils}[1]{\left\lceil#1\right\rceil}

\newcommand{\set}[1]{\left\{#1\right\}}

\newcommand{\T}{^\top}

\DeclareMathOperator*{\argmax}{arg\,max\,}
\DeclareMathOperator*{\argmin}{arg\,min\,}

\mathchardef\mhyphen="2D

\newcommand{\glmtsl}{\ensuremath{\tt GLM\mhyphen TSL}\xspace}

\newcommand{\lints}{\ensuremath{\tt LinTS}\xspace}
\newcommand{\linucb}{\ensuremath{\tt LinUCB}\xspace}

\newcommand{\ts}{\ensuremath{\tt TS}\xspace}
\newcommand{\ucb}{\ensuremath{\tt UCB1}\xspace}
\newcommand{\ucbglm}{\ensuremath{\tt UCB\mhyphen GLM}\xspace}

\newcommand{\sequentialhalving}{\ensuremath{\tt SH}\xspace}
\newcommand{\notes}{\ensuremath{\tt NoTeS}\xspace}
\newcommand{\uniform}{\ensuremath{\tt Uniform}\xspace}

\title{Empirical Bayes Regret Minimization}

\author{
  Chih-Wei Hsu \\
  Google Research \\
  \And
  Branislav Kveton \\
  Google Research \\
  \And
  Ofer Meshi \\
  Google Research \\
  \And
  Martin Mladenov \\
  Google Research \\
  \And
  Csaba Szepesv\'{a}ri \\
  DeepMind / University of Alberta
}

\begin{document}

\maketitle

\begin{abstract}
Most bandit algorithm designs are purely theoretical. Therefore, they have strong regret guarantees, but also are often too conservative in practice. In this work, we pioneer the idea of algorithm design by minimizing the empirical Bayes regret, the average regret over problem instances sampled from a known distribution. We focus on a tractable instance of this problem, the confidence interval and posterior width tuning, and propose an efficient algorithm for solving it. The tuning algorithm is analyzed and evaluated in multi-armed, linear, and generalized linear bandits. We report several-fold reductions in Bayes regret for state-of-the-art bandit algorithms, simply by optimizing over a small sample from a distribution.
\end{abstract}


\section{Introduction}
\label{sec:introduction}

A \emph{stochastic bandit} \cite{lai85asymptotically,auer02finitetime,lattimore19bandit} is an online learning problem where the \emph{learning agent} sequentially pulls arms with noisy rewards. The goal of the agent is to maximize its expected cumulative reward. Since the agent does not know the mean rewards of the arms in advance, it must learn them by pulling the arms. This results in the well-known \emph{exploration-exploitation trade-off}: \emph{explore}, and learn more about an arm; or \emph{exploit}, and pull the arm with the highest estimated reward thus far. In practice, the \emph{arm} may be a treatment in a clinical trial and its \emph{reward} is the outcome of that treatment on some patient population.

\emph{Optimism in the face of uncertainty} \cite{auer02finitetime,dani08stochastic,abbasi-yadkori11improved} and \emph{Thompson sampling (TS)} \cite{thompson33likelihood,agrawal12analysis} are arguably the most popular exploration designs in stochastic multi-armed bandits. In many problem classes, they match lower bounds, which indicates that the problems are \say{solved}. Unfortunately, these optimality results are typically \emph{worst-case}, in the sense that they hold for \emph{any problem instance} in the class. While this provides strong guarantees, it is not necessarily the best choice in practice. In this work, we focus on minimizing the \emph{Bayes regret}, an average regret over problem instances.

We assume that the learning agent has access to a distribution of problem instances, which can be sampled from. We propose to use these instances, as if they were simulators, to evaluate bandit algorithms. Then we choose the best empirical algorithm design. Our approach can be viewed as an instance of \emph{meta-learning}, or \emph{learning-to-learn} \cite{thrun96explanationbased,thrun98lifelong,baxter98theoretical,baxter00model}. It is also an application of \emph{empirical risk minimization} to learning what bandit algorithm to use. Therefore, we call it \emph{empirical Bayes regret minimization}. Finally, our approach can be viewed as an alternative to Thompson sampling, where we sample problem instances from the prior distribution and then choose the best algorithm design on average over these instances.

We make the following contributions in this paper. First, we propose \emph{empirical Bayes regret minimization}, an approximate minimization of the Bayes regret on sampled problem instances. Second, we propose two tractable instances of our problem, confidence interval and posterior width tuning. An interesting structure in these problems is that the regret is unimodal in the tunable parameter. Third, we propose a computationally and sample efficient algorithm, which we call \notes, that uses the unimodality to find a near-optimal tunable parameter. Fourth, we analyze \notes in the fixed-budget setting. Finally, we evaluate our methodology in multi-armed, linear, and generalized linear bandits. In all experiments, we observe significant gains that justify more empirical designs, as we argue for in this paper.


\section{Setting}
\label{sec:setting}

We start by introducing notation. The expectation operator is $\E{\cdot}$ and the corresponding probability measure is $\prob{\cdot}$. We define $[n] = \set{1, \dots, n}$ and denote the $i$-th component of vector $x$ by $x_i$.

A \emph{stochastic multi-armed bandit} \cite{lai85asymptotically,auer02finitetime,lattimore19bandit} is an online learning problem where the learning agent sequentially pulls $K$ arms in $n$ rounds. We formally define the problem as follows. Let $P$ be the joint probability distribution of arm rewards, with support $[0, 1]^K$. Let $(Y_t)_{t = 1}^n$ be a sequence of $n$ arm reward $K$-tuples $Y_t = (Y_{t, 1}, \dots, Y_{t, K})$, which are drawn i.i.d.\ from $P$. In round $t \in [n]$, the learning agent pulls arm $I_t \in [K]$ and observes its reward $Y_{t, I_t}$. The agent does not know $P$ or its mean, but can learn them from interactions.

The goal of the agent is to maximize its cumulative reward, which is equivalent to regret minimization. The \emph{$n$-round regret} of agent $A$ under distribution $P$ is defined as
\begin{align}
  R(A, P)
  = \sum_{t = 1}^n Y_{t, i_\ast} - \sum_{t = 1}^n Y_{t, I_t}\,,
  \label{eq:regret}
\end{align}
where $i_\ast = \argmax_{i \in [K]} \mu_i$ is the \emph{optimal arm} under distribution $P$, with mean $\mu = \condE{Y_1}{P}$; and $I_t $ denotes the pulled arm by agent $A$ in round $t$. The corresponding \emph{expected $n$-round regret} is $\condE{R(A, P)}{P}$, where the expectation is over the randomness in $(Y_t)_{t = 1}^n$ and potential randomness in $A$. Since $P$ fully characterizes the bandit problem, we refer to it as a \emph{problem instance}.

In this paper, we assume that the problem instance $P$ is drawn i.i.d.\ from a \emph{distribution of problem instances} $\cP$ and our goal is minimize the \emph{Bayes regret},
\begin{align}
  R(A)
  = \E{\condE{R(A, P)}{P}}\,,
  \label{eq:bayes regret}
\end{align}
where the outer expectation is over $P \sim \cP$. This makes our problem a variant of Bayesian bandits \cite{berry85bandit}. A celebrated solution to these problems is the Gittins index \cite{gittins79bandit}.

Note that the minimization of $R(A)$ is different from minimizing $\condE{R(A, P)}{P}$ for all $P$. While the latter is standard in multi-armed bandits, it is clearly more conservative because it optimizes equally for likely and unlikely problem instances $P$. The former is a good objective when the distribution $\cP$ can be estimated from past data. Such data are becoming increasingly available and are the main reason for recent works on off-policy evaluation from logged bandit feedback \cite{li11unbiased,swaminathan15counterfactual}.


\section{Empirical Bayes Regret Minimization}
\label{sec:empirical bayes regret minimization}

The key property of the Bayes regret, which we use in this work, is that it averages over problem instances $P \sim \cP$. By the tower rule of expectations, $R(A) = \E{\condE{R(A, P)}{P}} = \E{R(A, P)}$, where the last quantity can be minimized empirically when $\cP$ is known. In particular, let $P_1, \dots, P_s$ be $s$ i.i.d.\ samples from $\cP$ and $R(A, P_j)$ be the \emph{random regret} of agent $A$ on problem instance $P_j$, as defined in \eqref{eq:regret}. Let
\begin{align}
  \hat{R}(A)
  = \frac{1}{s} \sum_{j = 1} R(A, P_j)
\end{align}
be the \emph{empirical Bayes regret} of agent $A$. Then $\hat{R}(A) \to R(A)$ as $s \to \infty$.

The key idea in our work is to minimize $\hat{R}(A)$ instead of \eqref{eq:bayes regret}. Since $\hat{R}(A)$ is an average regret over $s$ runs of agent $A$ on $s$ problem instances, the minimization of $\hat{R}(A)$ over $A$ is equivalent to standard empirical risk minimization. It is also similar to meta-learning \cite{thrun96explanationbased,thrun98lifelong}, as learning of $A$ can be viewed as learning of an optimizer.

Note that $\hat{R}(A)$ can be minimized without having access to distributions $P_j$. In particular, it suffices to known all arm rewards, $Y^{(j)}_1, \dots, Y^{(j)}_n \sim P_j$. To see this, note that $\sum_{t = 1}^n Y_{t, i_\ast}$ in \eqref{eq:regret} does not depend on the agent. Thus the minimization of the Bayes regret is equivalent to maximizing the expected $n$-round reward. Similarly, the minimizers of $\hat{R}(A)$ are the maximizers of $\sum_{j = 1}^s \sum_{t = 1}^n Y_{t, I_t}^{(j)}$, the $n$-round reward across all problems. Therefore, our assumption that $\cP$ is known and can be sampled from is for simplicity of exposition.

\subsection{Confidence Width Tuning}
\label{sec:confidence width tuning}

Although the idea of minimizing $\hat{R}(A)$ over agents $A$ is conceptually simple, it is not clear how to implement it because the space of agents $A$ is hard to search efficiently. In this work, we focus on agents that are tuned variants of existing bandit algorithms. We explain this idea below on \ucb \cite{auer02finitetime}, a well-known bandit algorithm.

\ucb \cite{auer02finitetime} pulls the arm with the highest upper confidence bound (UCB). The UCB of arm $i \in [K]$ in round $t$ is
\begin{align}
  U_t(i)
  = \hat{\mu}_{t - 1, i} + \gamma \sqrt{2 \log(n) / T_{t - 1, i}}\,,
  \label{eq:ucb1}
\end{align}
where $\hat{\mu}_{t, i}$ is the average reward of arm $i$ in the first $t$ rounds, $T_{t, i}$ is the number of times that arm $i$ is pulled in the first $t$ rounds, and $\gamma = 1$. Roughly speaking, we propose choosing lower tuned $\gamma \in [0, 1]$ than $\gamma = 1$, which corresponds to the theoretically-sound design.

Tuning of $\gamma$ in \eqref{eq:ucb1} can be justified from several points of view. First, the confidence interval provides a natural trade-off between exploration and exploitation, as wider confidence intervals lead to more exploration. However, since it is usually designed by theory, it may ignore some structures in the bandit problem. Therefore, it is conservative and better empirical performance can be often achieved by scaling it down, by choosing $\gamma < 1$ in \eqref{eq:ucb1}. This practice is common in structured problems. For example, \citet{li10contextual} hand-picked $\gamma$, \citet{crammer11multiclass} tuned it using cross-validation, and \citet{gentile14online} tuned it on shorter horizons. We formally justify this practice.

Second, tuning of $\gamma$ in \eqref{eq:ucb1} is equivalent to tuning the probability $\delta$ that confidence intervals fail. In particular, for $\gamma = \sqrt{\log(1 / \delta) / \log(n)}$, the expected $n$-round regret of \ucb as a function of $\delta$ is $O(K \Delta^{-1} \log(1 / \delta) + (1 - \delta) n)$, where $\Delta$ is the minimum gap. This regret bound is convex in $\delta$. If the regret had a similar shape, and was for instance a unimodal quasi-convex function of $\gamma$, near-optimal values of $\gamma$ could be found efficiently by ternary search (\cref{sec:algorithm}). This is the first work that validates this trend empirically by large-scale simulations in many problems (\cref{sec:experiments}).

Finally, tuning of $\gamma$ in \eqref{eq:ucb1} can be also viewed as choosing the sub-Gaussian noise parameter of rewards. Specifically, \eqref{eq:ucb1} is derived for $\sigma^2$-sub-Gaussian rewards, where $\sigma^2 = 1 / 4$ is the maximum variance of a random variable on $[0, 1]$. If that variance was $1 / 16$, the confidence interval would be half the size and correspond to $\gamma = 1 / 2$.

\subsection{Posterior Width Tuning}
\label{sec:posterior width tuning}

Posterior width tuning is conceptually similar to confidence width tuning and also common in practice \cite{chapelle11empirical,zong16cascading}. We illustrate it below on Bernoulli Thompson sampling (\ts) \cite{agrawal12analysis}.

In Bernoulli \ts, the estimated mean rewards of arms are drawn from their posteriors and the arm with the highest mean is pulled. The posterior distribution of arm $i$ in round $t$ is $\mathrm{Beta}(\alpha_{t - 1, i}, \beta_{t - 1, i})$, where $\alpha_{t, i} = S_{t, i} + 1$, $\beta_{t, i} = T_{t, i} - S_{t, i} + 1$, $S_{t, i}$ is the cumulative reward of arm $i$ in the first $t$ rounds, and $T_{t, i}$ is the number of times that arm $i$ is pulled in the first $t$ rounds. Now note that $\var{\mathrm{Beta}(\alpha, \beta)} \approx \alpha \beta / (\alpha + \beta)^3$. Therefore, the standard deviation of the posterior is reduced by $\gamma$ when $\alpha$ and $\beta$ are divided by $\gamma^2$. We study this tuned variant of \ts in \cref{sec:experiments}.


\section{Sample-Efficient Tuning}
\label{sec:algorithm}

In this section, we propose an algorithm for minimizing the Bayes regret of bandit algorithms with a single tunable parameter, as in \cref{sec:confidence width tuning,sec:posterior width tuning}. We adopt the following notation. The space of tunable parameters is $\Gamma = [0, 1]$. The algorithm with tunable parameter $\gamma \in \Gamma$ is denoted by $A_\gamma$ and its Bayes regret is $R_\gamma = \E{R(A_\gamma, P)}$. The optimal tunable parameter is $\gamma_\ast = \argmin_{\gamma \in \Gamma} R_\gamma$.

Our algorithm outputs $\hat{\gamma} \in \Gamma$ and we maximize the probability that $\hat{\gamma}$ is \say{close} to $\gamma_\ast$, under a budget constraint on the number of measurements $s$ of the random regret. We focus on this fixed-budget setting because we want a practical algorithm for low budgets. The reason is that the estimation of the random regret is generally computationally costly, as it requires running a bandit algorithm.

Our design is motivated by ternary search, which is an iterative algorithm for minimizing unimodal functions, such as $R_\gamma$. The key idea in \emph{ternary search} is to reduce the hypothesis space by one third in each iteration. Specifically, let $[I_\ell, J_\ell] \subseteq \Gamma$ be the hypothesis space at the end of iteration $\ell \in [L]$. In iteration $\ell$, $[I_{\ell - 1}, J_{\ell - 1}]$ is divided by two points, $a < b$, into three intervals. If $R_a \geq R_b$, $\gamma_\ast$ cannot be in $[I_{\ell - 1}, a)$ and that interval is eliminated. Otherwise, $\gamma_\ast$ cannot be in $(b, J_{\ell - 1}]$ and that interval is eliminated. At the end of the last iteration $L$, ternary search outputs $\hat{\gamma} = (I_L + J_L) / 2$. The search is initialized by $I_0 = 0$ and $J_0 = 1$.

Because $R_\gamma$ is unknown, we propose \emph{noisy ternary search} (\notes), which searches for $\gamma_\ast$ based on the estimate of $R_\gamma$. The pseudocode of \notes is in \cref{alg:ternary ebrm}. The difference from ternary search is that the Bayes regret in each iteration $\ell$ is estimated from $s_\ell / 2$ problem instances (line \ref{alg:ternary ebrm:estimate}). \notes has two parameters, the number of iterations $L$ and budget allocation $(s_\ell)_{\ell = 1}^L$. In the next section, we show how to choose both to maximize the probability that $\hat{\gamma}$ is \say{close} to $\gamma_\ast$.

\begin{algorithm}[t]
  \caption{\notes: Noisy ternary search.}
  \label{alg:ternary ebrm}
  \begin{algorithmic}[1]
    \State \textbf{Inputs}: Number of iterations $L$, allocation $(s_\ell)_{\ell = 1}^L$
    \Statex
    \State $I_0 \gets 0, \, J_0 \gets 1$
    \For{$\ell = 1, \dots, L$}
      \State $\displaystyle
      a \gets \frac{2}{3} I_{\ell - 1} + \frac{1}{3} J_{\ell - 1}, \,
      b \gets \frac{1}{3} I_{\ell - 1} + \frac{2}{3} J_{\ell - 1}$
      \State Let $P_1, \dots, P_{s_\ell / 2}$ be i.i.d.\ samples from $\cP$
      \State $\displaystyle \hat{R}_a \gets
      \frac{2}{s_\ell} \sum_{k = 1}^{s_\ell / 2} R(A_a, P_k), \
      \hat{R}_b \gets
      \frac{2}{s_\ell} \sum_{k = 1}^{s_\ell / 2} R(A_b, P_k)$
      \label{alg:ternary ebrm:estimate}
      \If{$\hat{R}_a \geq \hat{R}_b$}
      $I_\ell \gets a, \, J_\ell \gets J_{\ell - 1}$
      \textbf{else}
      $I_\ell \gets I_{\ell - 1}, \, J_\ell \gets b$
      \EndIf
    \EndFor
    \State $\displaystyle \hat{\gamma} \gets (I_L + J_L) / 2$
  \end{algorithmic}
\end{algorithm}


\section{Analysis}
\label{sec:analysis}

We analyze $\notes$ in the fixed-budget setting. In particular, given budget allocation $(s_\ell)_{\ell = 1}^L$ and error tolerance $\eps \in [0, 1]$, we derive an upper bound on $\prob{|\hat{\gamma} - \gamma_\ast| > \eps}$, the probability that $\hat{\gamma}$ is not $\eps$-close to $\gamma$. This criterion is a natural continuous generalization of the probability of not identifying the best arm in fixed-budget best-arm identification \cite{karnin13almost,lattimore19bandit}. Our upper bound is given in \cref{thm:main} and we discuss how to minimize it in \cref{sec:discussion}. We make one assumption on $R_\gamma$.

\begin{assumption}
\label{ass:quasi-convex positive gradient} $R_\gamma$ is a unimodal quasi-convex function of $\gamma$ with minimum at $\gamma_\ast$ such that $\abs{R_a - R_b} \geq \lambda \abs{a - b}$ for all $a < b \leq \gamma_\ast$ and $a > b \geq \gamma_\ast$, for some $\lambda > 0$.
\end{assumption}

The above assumption is mild, and holds for any piecewise linear function where no segment is constant. Note that we do not assume anything when $a < \gamma_\ast < b$. In such cases, $\abs{a - b}$ can be large while $\abs{R_a - R_b}$ is small. We observe quasi-convexity empirically in all problems in \cref{sec:experiments}.

We also make a standard assumption for applying Hoeffding's inequality.

\begin{assumption}
\label{ass:sub-Gaussian noise} The random regret is $\sigma^2$-sub-Gaussian. That is, $\E{\exp[\lambda (R(A_\gamma, P) - R_\gamma)]} \leq \exp[\lambda^2 \sigma^2 / 2]$ holds for all $\gamma \in \Gamma$ and $\lambda > 0$.
\end{assumption}

We do not assume that $\sigma$ is known. Our analysis has two key steps. First, we derive the minimum number of steps to find an $\eps$-close solution.

\begin{lemma}
\label{lem:number of steps} Let $L \geq \log_{3 / 2}(1 /  (2 \eps))$. Then $J_L - I_L \leq 2 \eps$.
\end{lemma}

The lemma follows directly from the design of \notes, since $J_L - I_L = (2 / 3)^L = \exp[- L \log(3 / 2)]$. Now we bound the probability of event $\gamma_\ast \notin [I_L, J_L]$.

\begin{lemma}
\label{lem:error probability} We have $\displaystyle \prob{\gamma_\ast \notin [I_L, J_L]} \leq 4 \sum_{\ell = 1}^L \exp\left[- \frac{(2 / 3)^{2 \ell} \lambda^2 s_\ell}{32 \, \sigma^2}\right]$ for any number of iterations $L$ and $(s_\ell)_{\ell = 1}^L$.
\end{lemma}
\begin{proof}
First, we note that $\prob{\gamma_\ast \notin [I_L, J_L]} = \sum_{\ell = 1}^L \prob{E_\ell}$, where $E_\ell = \set{\gamma_\ast \in [I_{\ell - 1}, J_{\ell - 1}]\,, \gamma_\ast \notin [I_\ell, J_\ell]}$ is the event that $\gamma_\ast$ is eliminated incorrectly in iteration $\ell$. Now we bound each of the event probabilities below.

Fix iteration $\ell$. Let $i = I_{\ell - 1}$, $j = J_{\ell - 1}$, and $a$ and $b$ be defined as in \notes. Now note that the minimum $\gamma_\ast$ can be eliminated incorrectly in only two cases. The first case is $\hat{R}_a \leq \hat{R}_b$ and $\gamma_\ast \in (b, j]$. The second case is $\hat{R}_a \geq \hat{R}_b$ and $\gamma_\ast \in [i, a)$. Therefore, $\prob{E_\ell}$ decomposes as
\begin{align*}
  \prob{E_\ell}
  = \prob{\hat{R}_a \leq \hat{R}_b, \, \gamma_\ast \in (b, j]} +
  \prob{\hat{R}_a \geq \hat{R}_b, \, \gamma_\ast \in [i, a)}\,.
\end{align*}
We start with bounding $\prob{\hat{R}_a \leq \hat{R}_b, \, \gamma_\ast \in (b, j]}$. Let $\Delta = R_a - R_b$ and $m = R_a - \Delta / 2 = R_b + \Delta / 2$. From $a < b \leq \gamma_\ast$ and \cref{ass:quasi-convex positive gradient}, we have $R_a > R_b$ and that $\Delta > 0$. Thus, event $\hat{R}_a \leq \hat{R}_b$ can only occur if $\hat{R}_a \leq m$ or $\hat{R}_b \geq m$, which yields
\begin{align*}
  \prob{\hat{R}_a \leq \hat{R}_b, \, \gamma_\ast \in (b, j]}
  & \leq \prob{\hat{R}_a \leq m} + \prob{\hat{R}_b \geq m} \\
  & = \prob{R_a - \hat{R}_a \geq \Delta / 2} +
  \prob{\hat{R}_b  - R_b \geq \Delta / 2}\,.
\end{align*}
This is the sum of probabilities that $R_a$ and $R_b$ are estimated \say{incorrectly} by a large margin, at least $\Delta / 2$. We bound both probabilities using Hoeffding's inequality,
\begin{align*}
  \prob{R_a - \hat{R}_a \geq \Delta / 2} +
  \prob{\hat{R}_b  - R_b \geq \Delta / 2}
  \leq 2 \exp\left[- \frac{\Delta^2 s_\ell}{8 \sigma^2}\right]
  \leq 2 \exp\left[- \frac{(2 / 3)^{2 \ell} \lambda^2 s_\ell}{32 \, \sigma^2}\right]\,,
\end{align*}
where the last inequality is by using \cref{ass:quasi-convex positive gradient} in $\Delta = R_a - R_b \geq \lambda (b - a) = \lambda (j - i) / 3 = \lambda (2 / 3)^\ell / 2$. The upper bound on $\prob{\hat{R}_a \geq \hat{R}_b, \, \gamma_\ast \in [i, a)}$ is analogous. Now we chain all upper bounds on $\prob{E_\ell}$ and get our main claim.
\end{proof}

\noindent Finally, we take \cref{lem:number of steps,lem:error probability}, and note that for any $L$ in \cref{lem:number of steps}, $\prob{|\hat{\gamma} - \gamma_\ast| > \eps} \leq \prob{\gamma_\ast \notin [I_L, J_L]}$. This yields our main result.

\begin{theorem}
\label{thm:main} We have $\displaystyle \prob{|\hat{\gamma} - \gamma_\ast| > \eps} \leq 4 \sum_{\ell = 1}^L \exp\left[- \frac{(2 / 3)^{2 \ell} \lambda^2 s_\ell}{32 \, \sigma^2}\right]$ for any number of iterations $L$ in \cref{lem:number of steps} and $(s_\ell)_{\ell = 1}^L$.
\end{theorem}

\subsection{Discussion}
\label{sec:discussion}

We minimize the upper bound on $\prob{|\hat{\gamma} - \gamma_\ast| > \eps}$ in \cref{thm:main}, given a fixed integer budget $s > 0$, as follows. First, we set $L = \ceils{\log_{3 / 2}(1 /  (2 \eps))}$, which is the minimum permitted value by \cref{lem:number of steps}. Second, we select $s_\ell = c \, (3 / 2)^{2 \ell}$ for $\ell \in [L]$, where $c$ is a normalizer such that $s = \sum_{\ell = 1}^L s_\ell$. This budget allocation is not surprising. Since $J_\ell - I_\ell = (2 / 3)^\ell$, the elimination problem in iteration $\ell$ is $(3 / 2)^\ell$ harder than in iteration $1$, where $J_0 - I_0 = 1$. So we need $(3 / 2)^{2 \ell}$ more samples to attain the same error probability.


\section{Experiments}
\label{sec:experiments}

We experiment with various problems. In each experiment, we report the Bayes regret of all bandit algorithms as a function of their tunable parameters and show how \notes tunes it from a sample of the random regret. Each reported Bayes regret is estimated by the empirical Bayes regret from $10\,000$ i.i.d.\ samples from $\cP$. Note that this requires running a bandit algorithm $10\, 000$ times, for a single measurement, and is one of the largest empirical studies of bandit algorithms yet. For this reason, the reported standard errors of most measurements are very small.

\subsection{Warm-Up Experiment}
\label{sec:warm-up experiment}

Our first experiment is on Bernoulli bandits with $K = 2$ arms and horizon $n = 200$. The distribution of problem instances $\cP$ is over two problems, $\mu = (0.6, 0.4)$ and $\mu = (0.4, 0.6)$, each of which is chosen with probability $0.5$. Although this problem is simple, it already highlights the main benefits of our approach. The remaining experiments reconfirm them in more complex problems. We tune \ucb and Bernoulli \ts, as described in \cref{sec:posterior width tuning,sec:confidence width tuning}. Bernoulli \ts is chosen because it is near optimal in Bernoulli bandits \cite{agrawal12analysis}.

\begin{figure*}[t]
  \centering
  \includegraphics[width=1.8in]{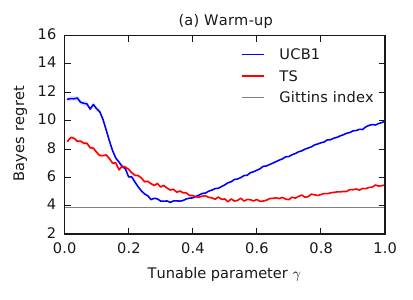}
  \includegraphics[width=1.8in]{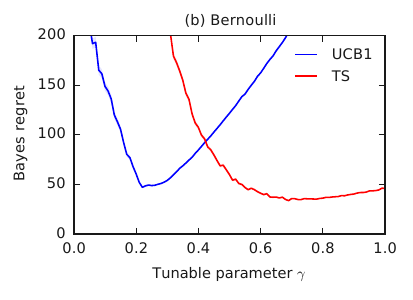}
  \includegraphics[width=1.8in]{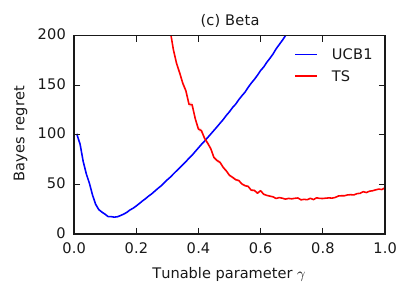}
  \vspace{-0.1in}
  \caption{The Bayes regret of tuned \ucb and \ts in three multi-armed bandit problems. The shaded areas are standard errors of the estimates.}
  \label{fig:multi-armed bandits}
\end{figure*}

In \cref{fig:multi-armed bandits}a, we show the Bayes regret of tuned \ucb and \ts as a function of the tunable parameter $\gamma$. The untuned theory-justified design corresponds to $\gamma = 1$. The regret is roughly unimodal. The minimum regret of \ucb is $4.2$ and is attained at $\gamma \approx 0.3$. This is a huge improvement over the regret of $10.0$ at $\gamma = 1$. The minimum regret of \ts is $4.2$, and is also lower than the regret of $5.5$ at $\gamma = 1$. In summary, we observe that both \ucb and \ts can be improved by tuning.

We tune \ucb and \ts by \notes for $\eps = 0.1$ and various budgets $s$. This setting of $\eps$ leads to learning reasonably good $\gamma$ in all experiments. The budget $s$ is allocated as suggested in \cref{sec:discussion}.

The Bayes regret after tuning by \notes is reported in \cref{tab:tuning}a. We observe the following trends. First, \ucb can be tuned well. Even at a low budget of $s = 50$, the regret of tuned \ucb is about a half of that of untuned \ucb. When $s = 1000$, the regret of tuned \ucb is comparable to that of the best design in hindsight. We observe similar trends for \ts, although the gains are not as significant.

\begin{table}[t]
  \centering
  {\small
  \begin{tabular}{llrr|rrr} \hline
    & & Best & Theory & \multicolumn{3}{c}{\notes} \\
    & Budget & & &
    $50$ & $200$ & $1000$ \\ \hline
    Warm-up & \ucb & $4.2$ & $10.0$ &
    $5.3$ & $4.7$ & $4.4$ \\
    & \ts & $4.3$ & $5.5$ &
    $5.1$ & $4.8$ & $4.6$ \\ \hline
    Bernoulli & \ucb & $47.2$ & $354.4$ &
    $63.6$ & $52.5$ & $49.2$ \\
    & \ts & $33.8$ & $46.1$ &
    $74.9$ & $42.9$ & $36.4$ \\ \hline
    Beta & \ucb & $17.2$ & $355.7$ &
    $19.3$ & $18.6$ & $17.8$ \\
    & \ts & $34.6$ & $46.8$ &
    $75.3$ & $42.5$ & $36.7$ \\ \hline
    Linear & \linucb & $49.1$ & $341.8$ &
    $76.2$ & $55.5$ & $50.6$ \\
    & \lints & $56.5$ & $341.4$ &
    $76.4$ & $60.2$ & $57.3$ \\ \hline
    Logistic & \ucbglm & $59.0$ & $193.7$ &
    $69.7$ & $63.4$ & $61.3$ \\
    & \glmtsl & $67.4$ & $377.8$ &
    $81.4$ & $72.2$ & $69.6$ \\ \hline
  \end{tabular}
  \quad
  \begin{tabular}{l|rrr}
    \multicolumn{4}{c}{} \\
    \multicolumn{4}{c}{} \\ \hline
    Budget &
    $200$ & $500$ & $1000$ \\ \hline
    & \multicolumn{3}{c}{\notes} \\
    \ucb & $26.9$ & $13.1$ & $4.8$ \\
    \ts & $60.8$ & $54.8$ & $50.1$ \\ \hline
    & \multicolumn{3}{c}{\uniform} \\
    \ucb & $30.0$ & $18.1$ & $12.9$ \\
    \ts & $69.7$ & $60.7$ & $53.5$ \\ \hline
    & \multicolumn{3}{c}{\sequentialhalving} \\
    \ucb & $23.0$ & $13.8$ & $9.4$ \\
    \ts & $64.6$ & $56.0$ & $51.9$ \\ \hline
  \end{tabular}
  } \\
  \vspace{0.05in} \hspace{1.1in} (a) \hspace{2.25 in} (b) \\
  \vspace{0.05in}
  \caption{\textbf{a.} The Bayes regret after tuning by \notes. The results are averaged over $10\,000$ runs. All standard errors are below $0.5$ and not reported to reduce clutter. \textbf{b.} Percentage of failures to find $\eps$-close solutions by \notes, \uniform, and \sequentialhalving. The results are averaged over $10\,000$ runs. All standard errors are below $0.5$ and not reported to reduce clutter. We do not show results for $s = 50$ because \sequentialhalving cannot be implemented in this setting.}
  \label{tab:tuning}
\end{table}

\subsection{Baselines}
\label{sec:baselines}

To assess the quality of tuned algorithms in \cref{sec:warm-up experiment}, we compute the Gittins index \cite{gittins79bandit}. This is an optimal dynamic-programming policy for Bayesian bandits. We compute it up to $200$ Bernoulli pulls, as described in Section 35 in \citet{lattimore19bandit}. This computation takes almost two days and requires roughly $200^8$ elementary operations. In comparison, our tuning takes seconds.

The Bayes regret of the Gittins index is $3.9$ and we show it as the gray line in \cref{fig:multi-armed bandits}a. Based on \cref{tab:tuning}a, \ucb and \ts can be tuned to a comparable regret. This shows that our tuning approach is reasonable, as it can attain a near-optimal regret. Unlike the Gittins index, it can be easily applied to structured problems (\cref{sec:linear experiment,sec:glm experiment}) and is more computationally efficient.

We also compare \notes to two best-arm identification algorithms that do not leverage the unimodal structure of our problem: uniform sampling (\uniform) and sequential halving (\sequentialhalving) \cite{karnin13almost}. In \uniform, the space of tunable parameters $\Gamma$ is discretized on an $\eps$-grid and each arm on the grid is allocated $\eps s$ samples. The arm with the lowest average regret is $\hat{\gamma}$. \sequentialhalving operates on the same $\eps$-grid and is run for $\ceils{\log_2 (1 / \eps)}$ iterations. In each iteration, the worst half of the arms is eliminated. The last remaining arm is $\hat{\gamma}$. The budget in iteration $\ell$ is $s_\ell \propto 2^\ell$. This allocation is similar to \notes and we choose it to make a fair comparison. Specifically, since only a $2^{- \ell}$ fraction of arms survives up to iteration $\ell$, each is allocated $2^{2 \ell}$ samples in that iteration.

The algorithms are compared in \cref{tab:tuning}b by our optimized criterion, $\prob{|\hat{\gamma} - \gamma_\ast| > \eps}$. When tuning \ucb, \notes outperforms \sequentialhalving at higher budgets. Both methods tune \ts comparably. The worst performing method is \uniform.

\subsection{Bernoulli Bandit}
\label{sec:bernoulli experiment}

This experiment is on Bernoulli bandits with $K = 10$ arms and horizon $n = 10\,000$. In comparison to \cref{sec:warm-up experiment}, we go beyond two arms and two problem instances in $\cP$. The distribution $\cP$ is defined as follows. The mean reward of arm $i$ is $\mu_i \sim \mathrm{Beta}(1, 1)$ for all $i \in [K]$. The rest of the setting is identical to \cref{sec:warm-up experiment}.

The Bayes regret of tuned \ucb and \ts is shown in \cref{fig:multi-armed bandits}b. The Bayes regret after tuning by \notes is reported in \cref{tab:tuning}a. We observe that both \ucb and \ts can be tuned well. Notably, even at a low budget of $s = 50$, tuned \ucb has five times lower regret than at $\gamma = 1$.

\subsection{Beta Bandit}
\label{sec:beta experiment}

This experiment is on beta bandits, where the distribution of arm $i$ is $\mathrm{Beta}(v \mu_i, v (1 - \mu_i))$ for $v = 4$, and the number of arms is $K = 10$. The parameter $v$ controls the maximum variance of rewards. The rest is identical to \cref{sec:bernoulli experiment}. The goal of this experiment is to study non-Bernoulli rewards.

We implement Bernoulli \ts with $[0, 1]$ rewards as suggested by \citet{agrawal12analysis}. For any reward $Y_{t, i} \in [0, 1]$, we draw pseudo-reward $\hat{Y}_{t, i} \sim \mathrm{Ber}(Y_{t, i})$ and then use it in \ts instead of $Y_{t, i}$. Although Bernoulli \ts can solve beta bandits, it is not statistically optimal anymore.

The Bayes regret of tuned \ucb and \ts is shown in \cref{fig:multi-armed bandits}c. Surprisingly, we observe that the minimum regret of \ucb, which is $17.2$, is lower than that of \ts, which is $34.6$. The reason is that in beta bandits, the variance of rewards is significantly lower than in Bernoulli bandits. Since Bernoulli \ts replaces beta rewards with Bernoulli rewards, it does not leverage this structure. On the other hand, tuning of $\gamma$ in \eqref{eq:ucb1} can be viewed as choosing the sub-Gaussian noise parameter of rewards, which leads to learning the structure.

The Bayes regret after tuning by \notes is reported in \cref{tab:tuning}a. We observe that both \ucb and \ts can be tuned well. Tuning of \ucb leads to better solutions than tuning of \ts, as discussed earlier. Even at a low budget of $s = 50$, tuned \ucb has more than $15$ times lower regret than at $\gamma = 1$.

\subsection{Linear Bandit}
\label{sec:linear experiment}

Linear bandits are arguably the simplest example of structured bandit problems and we investigate them in this experiment. In a linear bandit, the reward of arm $i$ in round $t$ is $Y_{t, i} = x_i\T \theta_\ast + \eps_{t, i}$, where $x_i \in \realset^d$ is a known feature vector of arm $i$, $\theta_\ast \in \realset^d$ is an unknown parameter vector that is shared by the arms, and $\eps_{t, i}$ is i.i.d.\ $\sigma^2$-sub-Gaussian noise. We set $\sigma = 0.5$. The distribution of problem instances $\cP$ is defined as follows. Each problem instance is a linear bandit with $K = 100$ arms and $d = 10$. Both $\theta_\ast$ and $x_i$ are drawn uniformly at random from $[-1, 1]^d$.

We tune \linucb \cite{abbasi-yadkori11improved}, which is a UCB algorithm, and \lints \cite{agrawal13thompson}, which is a posterior sampling algorithm. In \linucb, the UCB of arm $i$ in round $t$ is $U_t(i) = x_i\T \hat{\theta}_t + \gamma \, \sigma g(n) \sqrt{x_i\T G_t^{-1} x_i}$, where $\hat{\theta}_t$ is the ridge regression solution in round $t$, $G_t$ is the corresponding sample covariance matrix, and $g(n) = \tilde{O}(\sqrt{d})$ is a slowly growing function of $n$. The regularization parameter in ridge regression is $\lambda = 1$ and we set $g(n)$ as in \citet{abbasi-yadkori11improved}. In \lints, the value of arm $i$ in round $t$ is $x_i\T \tilde{\theta}_t$, where $\tilde{\theta}_t \sim \cN(\hat{\theta}_t, \gamma^2 \tilde{g}(n) G_t^{-1})$ is a sample from the posterior of $\theta_\ast$ scaled by a slowly growing function of $n$, $\tilde{g}(n) = \tilde{O}(d)$. We set $\tilde{g}(n)$ as in \citet{agrawal13thompson}.

\begin{wrapfigure}{r}{2.6in}
  \centering
  \vspace{-0.1in}
  \includegraphics[width=2.6in]{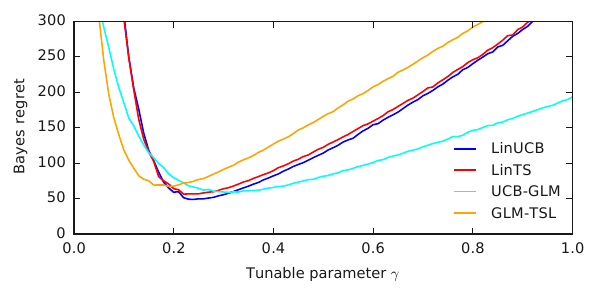}
  \caption{The Bayes regret of tuned UCB algorithms and posterior sampling in linear and GLM bandits. The shaded areas are standard errors of the estimates.}
  \label{fig:glm bandits}
  \vspace{-0.3in}
\end{wrapfigure}

The Bayes regret of tuned \linucb and \lints is shown in \cref{fig:glm bandits}. It is clearly unimodal and we observe that tuning can lead to huge gains. The Bayes regret after tuning by \notes is reported in \cref{tab:tuning}a. These results confirm that both algorithms can be tuned effectively. Even at a low budget of $s = 50$, the regret of tuned algorithms is four times lower than that of their untuned counterparts. At the highest budget of $s = 1000$, the regret approaches the minimum. We observe the same trends in generalized linear (GLM) bandits. This experiment is described in detail in \cref{sec:glm experiment}.


\section{Related Work}
\label{sec:background}

It is known that offline tuning of bandit algorithms tends to reduce their empirical regret \cite{vermorel05multiarmed,maes12metalearning,kuleshov14algorithms}. We go beyond these works in three aspects. First, we relate the problem to Bayesian bandits and empirical Bayes regret minimization. Second, we propose, analyze, and evaluate a sample-efficient tuning algorithm. Finally, we experiment with structured bandit problems, such as linear and GLM. Recently, \citet{duan16rl2} and \cite{boutilier20differentiable} proposed policy-gradient optimization of bandit policies. These approaches are not contextual and nowhere close to being as statistically efficient as \notes.

Our problem is an instance of meta-learning
\cite{thrun96explanationbased,thrun98lifelong}, where the objective is to learn a learning algorithm that performs well on future tasks, based on a sample of known tasks from the same distribution \cite{baxter98theoretical,baxter00model}. Recent years have seen a surge of interest in using meta-learning for deep reinforcement learning (RL) \cite{finn17modelagnostic,finn18probabilistic,mishra18simple}. Sequential multitask learning \cite{caruana97multitask} was studied in multi-armed bandits by \citet{azar13sequential} and in contextual bandits by \citet{deshmukh17multitask}. In comparison, our setting is offline. A general template for meta-learning of sequential strategies is presented in \citet{ortega19metalearning}. No efficient algorithms are provided.

Algorithm \notes is inspired by \citet{yu11unimodal}, who study the cumulative regret setting. In comparison, we study the fixed-budget best-arm identification setting, as in sequential halving \cite{karnin13almost}. We compare \notes to sequential halving in \cref{sec:baselines}.

Finally, there are many works on Bayesian bandits \cite{gittins79bandit,berry85bandit}, which focus on efficient Bayesian optimal methods. These methods require specific priors and are hard to apply to structured problems. In contrast, we do not make any strong assumption on the distribution of problem instances $\cP$ and our approach is more computationally efficient (\cref{sec:baselines}).


\section{Conclusions}
\label{sec:conclusions}

We propose empirical Bayes regret minimization, an approximate minimization of the Bayes regret of bandit algorithms on sampled problem instances. We justify this approach theoretically and evaluate it empirically. Our results show that empirical Bayes regret minimization leads to major gains in the empirical performance of existing bandit algorithms.

We leave open many questions of interest. First, we do not preclude that $\condE{R(A, P)}{P}$ can be minimized empirically jointly over all problem instances $P$. The main challenge is in the minimization over potentially infinite $\cP$ instead of averaging, which we do in this work. Second, one limitation of our work is that we optimize for a fixed horizon. Finally, we show that the Bayes regret is smooth in tunable parameters (\cref{fig:multi-armed bandits,fig:glm bandits}). This suggests that it can be optimized by gradient-based methods, which could be easily applied to multiple tunable parameters.

\bibliographystyle{plainnat}
\bibliography{References}

\clearpage
\onecolumn
\appendix


\section{Appendix}
\label{sec:appendix}

\subsection{Complete \cref{tab:tuning}}
\label{sec:complete table}

The complete version of \cref{tab:tuning} is in \cref{tab:all baselines,tab:all tuning}.

\begin{table}[t]
  \centering
  {\small
  \begin{tabular}{llrr|rrr|rrr|rrr} \hline
    & & Best & Theory &
    \multicolumn{3}{c|}{\notes} &
    \multicolumn{3}{c|}{\uniform} &
    \multicolumn{3}{c}{\sequentialhalving} \\
    & Budget & & &
    $50$ & $200$ & $1000$ & $50$ & $200$ & $1000$ & $50$ & $200$ & $1000$ \\ \hline
    Warm-up & \ucb & $4.2$ & $10.0$ &
    $5.3$ & $4.7$ & $4.4$ & $7.5$ & $5.1$ & $4.5$ & $5.6$ & $4.7$ & $4.4$ \\
    & \ts & $4.3$ & $5.5$ &
    $5.1$ & $4.8$ & $4.6$ & $6.5$ & $5.1$ & $4.6$ & $5.4$ & $4.8$ & $4.5$ \\
    Bernoulli & \ucb & $47.2$ & $354.4$ &
    $63.6$ & $52.5$ & $49.2$ & $137.0$ & $72.1$ & $54.2$ & $78.3$ & $55.4$ & $53.5$ \\
    & \ts & $33.8$ & $46.1$ &
    $74.9$ & $42.9$ & $36.4$ & $113.2$ & $43.9$ & $38.0$ & $55.2$ & $39.9$ & $37.0$ \\
    Beta & \ucb & $17.2$ & $355.7$ &
    $19.3$ & $18.6$ & $17.8$ & $52.6$ & $25.2$ & $18.2$ & $32.0$ & $19.0$ & $17.9$ \\
    & \ts & $34.6$ & $46.8$ &
    $75.3$ & $42.5$ & $36.7$ & $112.0$ & $44.1$ & $37.8$ & $55.5$ & $40.0$ & $36.5$ \\
    Linear & \linucb & $49.1$ & $341.8$ &
    $76.2$ & $55.5$ & $50.6$ & $146.3$ & $76.6$ & $59.0$ & $78.7$ & $59.5$ & $58.8$ \\
    & \lints & $56.5$ & $341.4$ &
    $76.4$ & $60.2$ & $57.3$ & $134.7$ & $75.1$ & $63.6$ & $80.1$ & $63.9$ & $63.7$ \\
    Logistic & \ucbglm & $59.0$ & $193.7$ &
    $69.7$ & $63.4$ & $61.3$ & $82.1$ & $66.0$ & $61.4$ & $70.5$ & $63.0$ & $59.6$ \\
    & \glmtsl & $67.4$ & $377.8$ &
    $81.4$ & $72.2$ & $69.6$ & $86.2$ & $76.9$ & $70.2$ & $86.1$ & $71.7$ & $69.6$ \\ 
    \hline
  \end{tabular}
  }
  \caption{The Bayes regret after tuning by \notes, \uniform, and \sequentialhalving. The results are averaged over $10\,000$ runs. All standard errors are below $0.5$ and not reported to reduce clutter.}
  \label{tab:all tuning}
\end{table}

\begin{table*}[t]
  \centering
  {\small
  \begin{tabular}{ll|rrr|rrr|rrr} \hline
    & &
    \multicolumn{3}{c|}{\notes} &
    \multicolumn{3}{c|}{\uniform} &
    \multicolumn{3}{c}{\sequentialhalving} \\
    & Budget &
    $200$ & $500$ & $1000$ & $200$ & $500$ & $1000$ & $200$ & $500$ & $1000$ \\ \hline
    Warm-up & \ucb &
    $26.9$ & $13.1$ & $4.8$ & $30.0$ & $18.1$ & $12.9$ & $23.0$ & $13.8$ & $9.4$ \\
    & \ts &
    $60.8$ & $54.8$ & $50.1$ & $69.7$ & $60.7$ & $53.5$ & $64.6$ & $56.0$ & $51.9$ \\
    Bernoulli & \ucb &
    $7.5$ & $2.3$ & $0.2$ & $17.8$ & $4.0$ & $0.5$ & $1.5$ & $0.1$ & $0.1$ \\
    & \ts &
    $45.6$ & $29.1$ & $22.5$ & $48.4$ & $39.7$ & $33.3$ & $43.1$ & $36.3$ & $34.0$ \\
    Beta & \ucb &
    $0.4$ & $0.2$ & $0.0$ & $7.5$ & $1.4$ & $0.3$ & $0.7$ & $0.2$ & $0.0$ \\
    & \ts &
    $44.8$ & $30.9$ & $23.2$ & $47.7$ & $36.9$ & $31.6$ & $42.2$ & $33.1$ & $29.9$ \\
    Linear & \linucb &
    $5.6$ & $2.1$ & $0.5$ & $9.3$ & $1.5$ & $0.7$ & $1.4$ & $0.8$ & $0.7$ \\
    & \lints &
    $4.3$ & $1.8$ & $0.2$ & $7.0$ & $1.4$ & $0.8$ & $1.3$ & $0.6$ & $0.8$ \\
    Logistic & \ucbglm &
    $12.8$ & $2.8$ & $0.6$ & $23.3$ & $15.9$ & $9.4$ & $17.1$ & $7.8$ & $2.8$ \\
    & \glmtsl &
    $0.5$ & $0.0$ & $0.0$ & $6.0$ & $1.8$ & $0.2$ & $4.0$ & $0.6$ & $0.1$ \\ \hline
  \end{tabular}
  }
  \caption{Percentage of failures to find $\eps$-close solutions by \notes, \uniform, and \sequentialhalving. The results are averaged over $10\,000$ runs. All standard errors are below $0.5$ and not reported to reduce clutter. We do not show results for $s = 50$ because \sequentialhalving cannot be implemented in this setting.}
  \label{tab:all baselines}
\end{table*}

\subsection{GLM Bandit}
\label{sec:glm experiment}

Generalized linear (GLM) bandits are another class of structured bandit problems. We focus on logistic bandits, with binary rewards. The reward of arm $i$ in round $t$ is $Y_{t, i} = \mu(x_i\T \theta_\ast) + \eps_{t, i}$, where $x_i \in \realset^d$ is a known feature vector of arm $i$, $\theta_\ast \in \realset^d$ is an unknown parameter vector that is shared by the arms, $\mu(v)$ is a sigmoid function, and $\eps_{t, i}$ is i.i.d.\ $\sigma^2$-sub-Gaussian noise. The distribution of problem instances $\cP$, over $x_i$ and $\theta_\ast$, is the same as in \cref{sec:linear experiment}.

We tune \ucbglm \cite{li17provably}, which is a UCB algorithm, and \glmtsl \cite{abeille17linear}, which is a posterior sampling algorithm. In \ucbglm, the UCB of arm $i$ in round $t$ is $U_t(i) = x_i\T \hat{\theta}_t + \gamma \, \frac{\sigma}{\kappa} g(n) \sqrt{x_i\T G_t^{-1} x_i}$, where $\hat{\theta}_t$ is the maximum likelihood estimate (MLE) of $\theta_\ast$ in round $t$, $G_t$ is the sample covariance matrix in \cref{sec:linear experiment}, and $g(n) = \tilde{O}(\sqrt{d})$ is a slowly growing function of $n$. We choose $g(n)$ as in the analysis of \citet{li17provably}. We set $\kappa$ to $0.25$, which is the maximum derivative of $\mu$, and thus the most optimistic setting. In \glmtsl, the value of arm $i$ in round $t$ is $x_i\T \tilde{\theta}_t$, where $\tilde{\theta}_t \sim \cN(\hat{\theta}_t, \gamma^2 \tilde{g}(n) H_t^{-1})$ is a sample from the Laplace approximation to the posterior distribution of $\theta_\ast$, $\tilde{g}(n) = \tilde{O}(d)$ is a slowly growing function of $n$, and $H_t$ is a weighted sample covariance matrix. We set $\tilde{g}(n)$ as in the analysis of \citet{kveton19randomized}.

The Bayes regret of tuned \ucbglm and \glmtsl is shown in \cref{fig:glm bandits}. Similarly to linear bandits, it is clearly unimodal and we observe that tuning can lead to huge gains. The Bayes regret after tuning by \notes is reported in \cref{tab:tuning}a. As in linear bandits, tuning leads to significant gains.

\end{document}